\documentclass{bmvc2k}
\usepackage{wrapfig}
\usepackage{graphicx}
\usepackage{amsmath}
\usepackage{amssymb}
\usepackage{booktabs} 
\title{Depth-to-Image Synthesis-Driven Generative Unguided Depth Completion}

\addauthor{Jiayi Yuan}{jiayi_yuan@mymail.sutd.edu.sg}{1}
\addauthor{Na Zhao$^*$}{na_zhao@sutd.edu.sg}{1}
\addauthor{De Wen Soh}{dewen_soh@sutd.edu.sg}{1}

\addinstitution{
 Singapore University of Technology and Design\\
 Singapore
}

\runninghead{Jiayi Yuan, Na Zhao, De Wen Soh}{GUDC}

\def\eg{\emph{e.g}\bmvaOneDot}
\def\ie{\emph{i.e}\bmvaOneDot}

\def\etal{\emph{et al}\bmvaOneDot}

\begin{document}

\maketitle

\begin{abstract}
Guided depth completion methods heavily depend on RGB quality and alignment, while unguided ones often suffer from limited precision due to the absence of explicit visual cues. In this paper, we present Depth-to-Image Synthesis-Driven Generative Unguided Depth Completion \textbf{(GUDC)}, a new completion paradigm that innovatively bridges advanced 2D generative models with unguided depth completion, enabling semantics-aware depth inference without real RGB inputs. Our key idea is to exploit ControlNet’s powerful depth-conditioned generation capability to synthesize pseudo-images directly from sparse depth, effectively converting the original unguided setting into a semantics-guided one. 
To address the potential image-depth misalignment caused by depth sparsity, we propose a multi-level dense-to-sparse representation distillation strategy for ControlNet fine-tuning, where dense-depth features act as teacher signals to distill consistent structural representations for sparse-depth inputs. Furthermore, during pseudo-image-guided completion, we propose a pseudo-image semantic attention fusion module to adaptively extract informative semantic cues from pseudo-images while suppressing artifacts (\eg, texture hallucinations). Extensive experiments on KITTI and NYUv2 validate that our GUDC achieves superior accuracy and robustness over existing methods.
\end{abstract}

%-------------------------------------------------------------------------
\section{Introduction}
\label{sec:intro}
Depth completion, which aims to recover dense depth maps from sparse measurements, is a fundamental task in 3D scene understanding, with broad applications in autonomous driving~\cite{you2019pseudo, song2021self}, indoor robotics~\cite{newcombe2011kinectfusion, teixeira2020aerial}, and 3D reconstruction~\cite{Yuan_2026_CVPR}. 
Recent advances have been driven by leveraging aligned RGB images as visual guidance to restore fine-grained structures and boundaries from sparse depth inputs. However, such image-guided approaches are fundamentally limited by the availability and quality of RGB inputs, which may be unreliable in many practical scenarios, such as nighttime driving, low-light environments, or LiDAR-only sensing systems. This limitation motivates the study of unguided depth completion, where the corresponding RGB images are not available during inference.

\begin{figure}[t]
  \centering
  \includegraphics[width=0.85\linewidth]{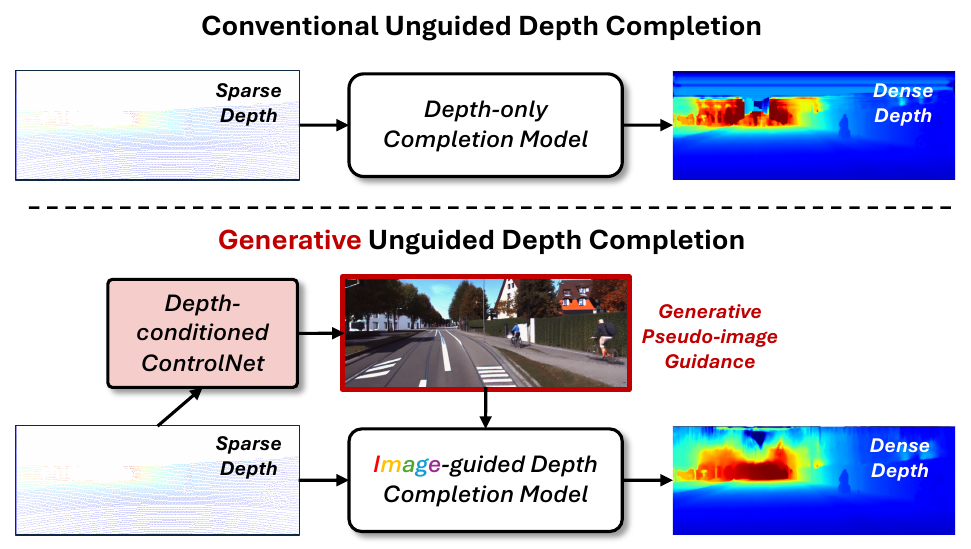}
  \caption{Comparison of conventional unguided depth completion and our proposed generative unguided depth completion framework (GUDC). Unlike prior methods that operate solely on sparse depth, our approach employs a depth-conditioned ControlNet to synthesize geometrically aligned pseudo-images, providing semantic priors and structural visual cues that effectively guide the completion process.
}
\label{intro_fig}
\end{figure}

In the context of unguided depth completion, deep learning-based methods can be broadly categorized into \textit{sparsity-aware} approaches and \textit{implicit auxiliary image-guided} approaches. Sparsity-aware approaches \cite{uhrig2017sparsity, jaritz2018sparse, huang2019hms, eldesokey2020uncertainty} typically employ binary masks or continuous confidence maps to identify reliable depth observations and restrict feature aggregation to valid measurements.  However, due to the absence of visual guidance, these methods often struggle to recover object boundaries and structural details, especially in severely sparse regions. 
To compensate for the lack of semantic context, implicit auxiliary image-based approaches~\cite{lu2020depth,lu2022depth} introduce an additional image branch during training that either reconstructs grayscale images~\cite{lu2020depth} or guides latent depth features using an image-based supervision loss~\cite{lu2022depth}. While these methods provide implicit image-level cues, they often fail to reconstruct meaningful content in unseen scenes and lack explicit, fine-grained visual guidance at inference time, thereby limiting their effectiveness and generalization capabilities. Moreover, the introduction of auxiliary tasks may lead to optimization conflicts and typically fails to yield controllable and transferable semantic representations.

Motivated by the recent success of generative AI models~\cite{ho2020denoising, rombach2022high, yang2023diffusion}, we introduce \textit{Depth-to-Image Synthesis-Driven Generative Unguided Depth Completion \textbf{(GUDC)}}, a new completion paradigm that bridges advanced 2D generative modeling with unguided depth completion to enhance completion accuracy.
As shown in Fig.~\ref{intro_fig}, our core idea is to synthesize pseudo-images that are geometrically consistent with the sparse depth input while encoding fine-grained structural details and scene-level semantic information.
These pseudo-images serve as explicit and reliable visual guidance, transforming the original unguided task into a semantics-guided one.
In contrast to implicit auxiliary image-based methods, our GUDC provides semantic and structural guidance in an explicit manner during both training and inference phases, leading to more accurate, robust, and interpretable depth completion.

Specifically, inspired by the powerful depth-conditioned image synthesis capability of ControlNet~\cite{zhang2023adding}, we first employ it to generate pseudo-images conditioned on the given sparse depth inputs.
However, due to the inherent sparsity and non-uniform distribution of depth measurements, directly applying a pre-trained ControlNet may lead to suboptimal alignment between the generated images and the input depth.  
To address this, we propose a \textit{multi-level dense-to-sparse representation distillation} strategy for ControlNet fine-tuning, enhancing both image-depth consistency and generative fidelity under sparse conditions.
Specifically, features extracted from dense ground-truth depth maps act as teacher signals to supervise the corresponding sparse-depth representations, enforcing geometry-aware structural consistency throughout the distillation process.

In the pseudo-image-guided completion phase, to mitigate potential uncertainties in the pseudo-images (\eg, texture hallucinations and spurious edges), we introduce a \textit{pseudo-image semantic attention fusion} (PSAF) module to enable more reliable semantic enhancement of depth. Specifically, PSAF first extracts intermediate visual features from the diffusion process, which contain rich contextual information and stable semantic cues. Guided by these semantic anchors, PSAF then applies a semantic-guided affinity kernel and cross-modality attention to adaptively retrieve key features from the pseudo-images, %for 
yielding enhanced depth representation.
It is worth noting that our generative unguided depth completion framework is \textit{model-agnostic} and can be seamlessly integrated into a wide range of existing guided depth completion architectures.

To summarize, our main contributions are as follows:

\begin{itemize}
    \item We are \textit{the first} to introduce a generative unguided depth completion paradigm that employs depth-conditioned diffusion models to generate pseudo-images as explicit visual guidance, effectively transforming the original unguided task into a semantically guided one for improved completion quality.
    \item To enhance image generation quality, we propose a multi-level dense-to-sparse representation distillation strategy that fine-tunes ControlNet by aligning sparse-depth representations with dense-depth teacher features, thereby improving structural consistency and generation fidelity under highly sparse conditions.
  \item To mitigate inherent uncertainties in pseudo-images, we propose a pseudo-image semantic attention fusion module that leverages diffusion features as stable semantic anchors to guide affinity-based image feature correction and cross-modality visual-geometric fusion for refined depth estimation.
  \item Our framework is general and plug-and-play, seamlessly integrating with diverse depth completion backbones. Extensive experiments on indoor and outdoor benchmarks demonstrate its effectiveness and robustness in challenging scenarios.
    
\end{itemize}

\section{Related Work}

\noindent\textbf{Guided Depth Completion.}
Guided depth completion methods rely on aligned RGB images to provide structural and semantic context for recovering dense depth from sparse measurements. Early approaches adopt classical image processing techniques \cite{ku2018defense} (\eg, dilation and hole filling) and joint bilateral filtering \cite{qi2013structure, chen2012depth, richardt2012coherent} to propagate sparse depth values. However, these methods often suffer from over-smoothing and poor generalization in complex scenes.
With the advent of deep learning, subsequent works leverage convolutional neural networks to learn effective RGB-depth fusion strategies. A common paradigm \cite{jaritz2018sparse, shivakumar2019dfusenet, yuan2023structure, li2020multi, tang2024bilateral, yuan2023recurrent, park2024test, zhu2025svdc} employs dual-branch encoders to extract features from sparse depth maps and corresponding RGB images, followed by feature fusion via a shared decoder. For instance, Li \etal~\cite{li2020multi} proposed a cascaded hourglass network with one image branch and three scale-specific depth branches to enhance multi-scale feature extraction. To mitigate over-smoothing, Park \etal~\cite{park2020non} introduced a post-processing stage based on propagation networks that iteratively refine the initial depth predictions. Tang \etal~\cite{tang2024bilateral} further addressed ambiguity and scale sensitivity by proposing a three-stage pipeline to support downstream fusion.
Despite strong performance, these methods heavily depend on the calibration accuracy and visual quality of RGB inputs, limiting their applicability in real-world scenarios where images may be noisy, misaligned, or unavailable.

\noindent\textbf{Unguided Depth Completion.} Given a sparse depth map, unguided depth completion methods aim to directly predict dense depth without relying on corresponding RGB images. 
Early works \cite{uhrig2017sparsity, chodosh2019deep, huang2019hms, jaritz2018sparse} design sparsity-aware CNNs by introducing binary validity masks to distinguish between observed and missing pixels during convolution, enabling standard networks to better handle sparse inputs. However, these masks tend to saturate in the early convolutional layers, leading to degraded performance in deeper layers \cite{jaritz2018sparse, eldesokey2019confidence}. 
To address this issue, Eldesokey \etal~\cite{eldesokey2018propagating} proposed the normalized convolutional neural network that replaces binary masks with continuous confidence maps for uncertainty-aware filtering. A follow-up work \cite{eldesokey2020uncertainty} further introduced a self-supervised learning scheme to estimate the input confidence maps and suppress noisy measurements. 
Nevertheless, these purely geometric methods often struggle in complex scenes due to the lack of visual semantics and structural priors. To alleviate this, Lu \etal~\cite{lu2020depth} introduced an auxiliary RGB reconstruction branch during training to inject visual semantics into the depth estimation pipeline. Similarly, Lu \etal~\cite{lu2022depth} employed a latent-space autoencoder to generate image features, which are then used to regress dense depth. 
While these methods improve generalization by incorporating visual priors, they still lack explicit visual guidance at inference time, limiting their ability to recover fine-grained structures and object boundaries. 
In contrast, our approach introduces explicit and reusable visual cues at inference by generating pseudo-images that are both semantically informative and geometrically aligned with the sparse depth. 

\noindent\textbf{Diffusion Models for Visual Guidance.} 
Recent advances in generative models, particularly diffusion models \cite{ho2020denoising, rombach2022high}, have shown remarkable capabilities in synthesizing photorealistic images with fine-grained structures. ControlNet \cite{zhang2023adding} further enhances these models by introducing spatially aligned conditional control, enabling precise image generation from structured inputs such as sparse depth maps, edges, or segmentation masks. Several studies have explored diffusion models for scene-level synthesis \cite{poole2022dreamfusion, liu2023zero} and image-conditioned depth estimation \cite{tosi2024diffusion}. Notably, Wang \etal~\cite{wang2024freereg} investigated the intermediate features extracted from depth-to-image diffusion models and found them to be semantically consistent between images and point clouds, enabling applications like image-to-point cloud registration.
Despite these advancements, the potential of generative models for unguided depth completion remains largely unexplored, particularly as a means of compensating for the absence of RGB inputs by providing semantically enriched guidance. Here, we are the first to propose a novel generative unguided depth completion framework that employs a depth-conditioned ControlNet to generate pseudo-images. Such pseudo-images are geometrically aligned with the sparse depth input, effectively converting the original unguided setting into a semantically guided one. Moreover, to mitigate the potential uncertainty introduced by generative artifacts (\eg, texture hallucinations), we extract diffusion features as stable semantic cues to guide more reliable image-depth feature fusion.

\begin{figure}[t]
  \centering
\includegraphics[width=1\linewidth]{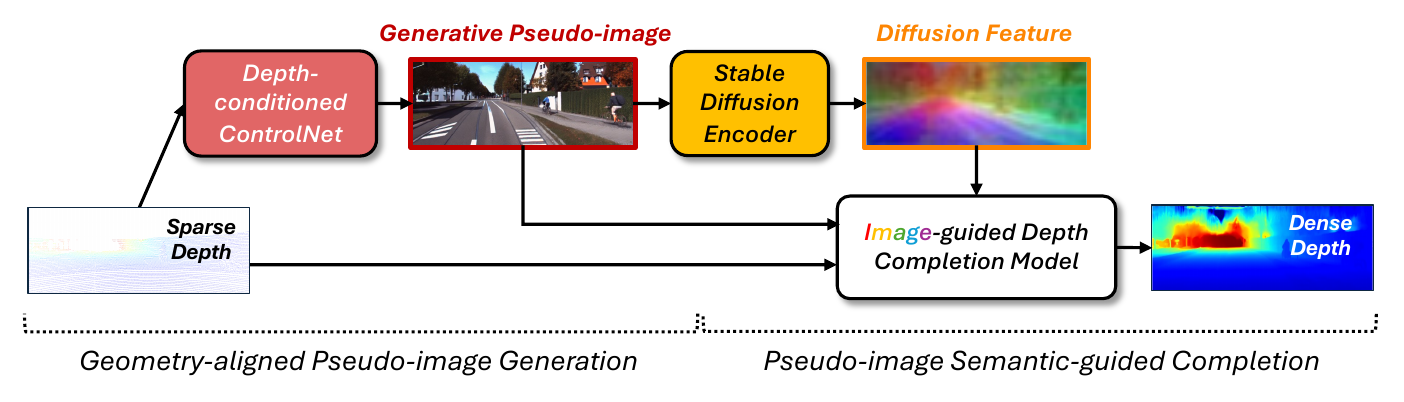}
  \caption{\textit{Framework of our proposed Generative Unguided Depth Completion}.
 Given sparse depth, we first apply a depth-conditioned ControlNet to generate a geometrically consistent and high-quality pseudo-image. This pseudo-image serves as structural and contextual guidance for subsequent image-guided depth completion. To mitigate potential uncertainties in the generated image, Stable Diffusion is employed as a semantic encoder to extract robust semantic cues from the pseudo-image. These cues are fused with the sparse depth and pseudo-image within an image-guided completion network to predict the final dense depth.
}
\label{pipeline}
\end{figure}

\section{Method}

\textbf{Problem Definition.}
Given a sparse depth map \( D_s \in \mathbb{R}^{H \times W} \), the objective of depth completion is to predict a dense depth map \( \hat{D} \in \mathbb{R}^{H \times W} \) that accurately fills in the missing depth values in \( D_s \).  
This task is typically formulated as learning a mapping function from sparse and incomplete depth observations to a dense depth representation, which can be modeled by a deep neural network parameterized by \( \theta \) as follows:
\begin{equation}
\hat{D} = \mathcal{F}(D_s; \theta).
\label{eq:unguided}
\end{equation}

In the \textit{unguided setting}, the model operates solely on the sparse input without access to image guidance, as shown in Eq.~\ref{eq:unguided}. 
By contrast, \textit{image-guided} methods typically assume that the corresponding RGB image \( I \in \mathbb{R}^{H \times W \times 3} \) is accessible to enhance depth completion through image-depth feature interactions. This paradigm can be generally formulated as:
\begin{equation}
\hat{D} = \mathcal{F}(D_s, I; \theta).
\end{equation}

In our setting, no real RGB image is available at inference time. Instead, we generate a pseudo-image
$\tilde{I}=G(D_s)$ from the sparse depth itself and formulate the completion as:
\begin{equation}
\hat{D}=\mathcal{F}(D_s,\tilde{I};\theta).
\end{equation}

\subsection{Overview}
On one hand, recent image-guided depth completion methods have shown that incorporating RGB images can significantly enhance completion quality by providing rich semantic context and fine-grained visual cues. On the other hand, RGB images are not always accessible or reliable during inference in practice. As a result, unguided depth completion using only sparse depth measurements is more practical, but it often yields suboptimal completion precision due to the absence of an image. To strike a balance between accuracy and practicality, we propose \textit{Generative Unguided Depth Completion (GUDC)}, a novel unguided framework that generates high-quality pseudo-images from sparse depth inputs. Consequently, we transform the original unguided setting into a semantically guided one, as illustrated in Fig.~\ref{pipeline}.

To ensure that the generated pseudo-images provide high-quality and geometry-consistent completion guidance, our framework is built upon two key components:
\textbf{i) Geometry-aligned pseudo-image generation:} To preserve precise image-depth alignment as well as achieve high visual fidelity, we introduce a novel geometry-aligned pseudo-image generation strategy based on a fine-tuned, depth-conditioned ControlNet (\textcolor{blue}{Sec.~\ref{img_gen}});
\textbf{ii) Pseudo-image semantic-guided depth completion:} In the pseudo-image-guided completion phase, to mitigate potential uncertainties in the generated pseudo-images, we introduce a pseudo-image semantic attention fusion module that adaptively emphasizes reliable regions and enables more robust and semantically enriched depth representation learning (\textcolor{blue}{Sec.~\ref{dep_enh}}).

\subsection{Geometry-aligned Pseudo-image Generation}
\label{img_gen}
Inspired by the powerful conditional generation capability of ControlNet, a controllable extension of Stable Diffusion, we leverage it as a generative prior to synthesize geometry-consistent pseudo-images conditioned on sparse depth maps.

\noindent\textbf{Stable Diffusion.}
\textit{Stable Diffusion}~\cite{rombach2022high} is a latent text-to-image diffusion model that generates images by iteratively denoising latent representations, conditioned on a text prompt.
It performs denoising within a compressed latent space learned by a pretrained autoencoder, significantly improving computational efficiency and memory usage.
The model refines a noisy latent variable \( x_t \) using a denoiser \( \epsilon_\theta(x_t; t, c) \), where \( t \) denotes the diffusion timestep and \( c \) is the token sequence of the text prompt.
Here, the denoiser is implemented as a UNet-like architecture that incorporates both self-attention and cross-attention blocks.

\begin{wrapfigure}{r}{0.5\textwidth}
  \centering
    \vspace{-3.5mm}
  \includegraphics[width=\linewidth]{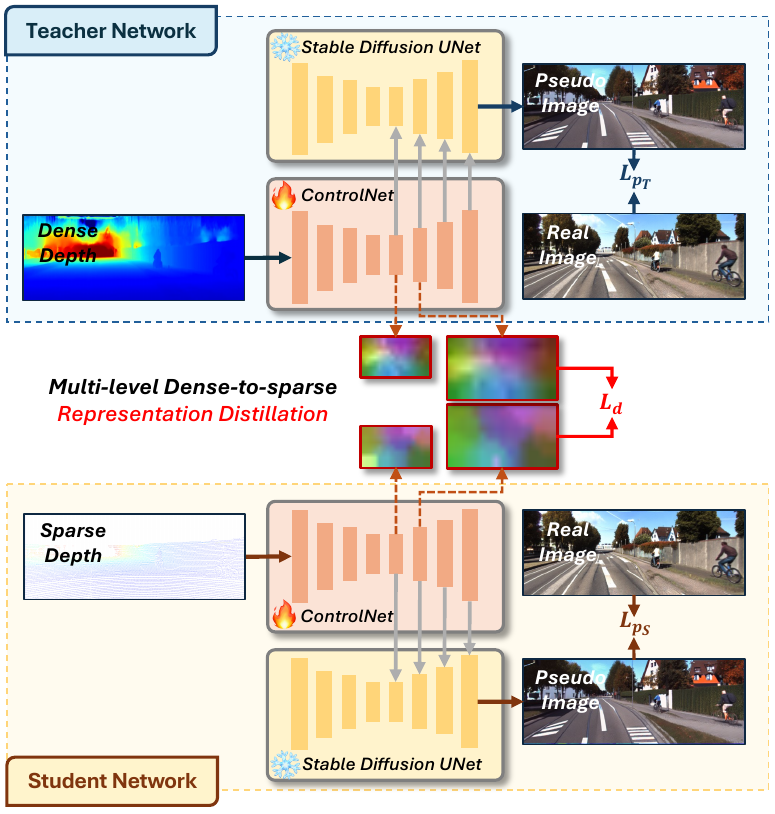}
  \vspace{-2mm}
  \caption{\textit{Pipeline of our multi-level dense-to-sparse representation distillation}, which transfers structural representations from a dense-depth teacher to a sparse-depth student.}
  \label{Distillation}
\end{wrapfigure}
\noindent\textbf{ControlNet-driven Pseudo-Image Generation.}
Beyond text-conditioned generation, ControlNet~\cite{zhang2023adding} extends Stable Diffusion by incorporating spatial conditioning inputs such as edges, depth maps, or segmentation masks, enabling fine-grained structural control over the generation process.
We adopt its depth-conditioned variant to synthesize geometry-consistent pseudo-images from sparse depth maps, providing explicit visual guidance for the subsequent depth completion stage.
To effectively integrate depth conditions, ControlNet introduces a trainable copy of the pretrained Stable Diffusion encoder as an auxiliary branch, enabling depth encoding while preserving the original representational capacity.
Depth features ${\mathbf{f}_i^{dpt}}$ and noisy image features ${\mathbf{f}_i^{img}}$ are extracted independently and progressively fused within the Stable Diffusion U-Net decoder (\ie, the denoiser).
Formally, the input $g_j$ to the $j$-th decoder layer is defined as:
\begin{equation}
g_j = \operatorname{concat}(g_{j-1}, \mathbf{f}_i^{img} + \operatorname{zero}(\mathbf{f}_i^{dpt})),
\end{equation}
where $i+j=L$ and $L=12$ denotes the total number of decoding layers in the Stable-Diffusion UNet. The term $\text{zero}(\cdot)$ refers to zero-initialized convolutional layers introduced by ControlNet, which initially suppress the conditional features' influence during early training phases. This design allows gradual learning of the conditional branch without adversely impacting the original generation capabilities. 

\noindent\textbf{Multi-level Dense-to-sparse Distillation-based ControlNet Fine-tuning.} 
Although pretrained ControlNet performs well with dense depth conditions, its direct application to sparse depth conditioning often yields poor geometric consistency and degraded image quality.
Naive fine-tuning also fails to impose stable geometric constraints, as the sparse depth provides insufficient supervisory signals to guide the model’s internal representations.
This leads to representation instability and weak depth-image correspondence, ultimately causing structural distortions and missing details in the generated pseudo-images (Fig.~\ref{ab_results}$(a)$).

To address the aforementioned limitations, we introduce a \textit{multi-level dense-to-sparse representation distillation} strategy for ControlNet fine-tuning. Our core idea is to explicitly transfer structural priors from dense-depth representations to sparse-depth ones, enabling ControlNet to encode geometry-aware knowledge even under highly incomplete conditions.
Specifically, as illustrated in Fig.~\ref{Distillation}, we first pre-train a teacher ControlNet $\Phi_\mathcal{T}$ on dense ground-truth depth maps $\mathcal{D}_{d}$ to learn structurally informative representations for geometry-consistent image generation. The teacher is optimized using the standard denoising objective as follows: 
\begin{equation}
\mathcal{L}_{p_{\mathcal{T}}} = \mathbb{E}_{{x}_t, t, \tilde{c}, {d}_d, \epsilon \sim \mathcal{N}(0, 1)} \left[\|\epsilon - \tilde{\epsilon}_\theta({x}_t, t, \tilde{c}, {d}_d)\|_2^2\right],
\end{equation}
where ${x}_t, {d}_d$ represent the diffused latent representations of the image $\mathcal{I}$ and dense depth map $\mathcal{D}_d$, respectively, and $\tilde{c}$ is the conditioning token sequence derived from the text prompt.

Once trained, the teacher model is frozen and employed to extract multi-level ControlNet feature representations ${f_{\Phi_\mathcal{T}}^{(l)}}$ conditioned on the dense depth input $\mathcal{D}_d$, where $l$ denotes the layer index.
A separate student ControlNet $\Phi_\mathcal{S}$, conditioned on sparse depth maps $\mathcal{D}_s$, is then optimized to align its internal multi-level features ${f^{(l)}_{\Phi_\mathcal{S}}}$ with those of the teacher via a layer-wise distillation loss as follows:
\begin{equation}
\mathcal{L}_{d} = {\textstyle \sum_{l=1}^{L}} \left\| f^{(l)}_{\Phi_\mathcal{S}} - \operatorname{SG}(f^{(l)}_{\Phi_\mathcal{T}}) \right\|_2^2,
\end{equation}  
where $\operatorname{SG}(\cdot)$ indicates the stop-gradient operation that prevents backpropagation through the teacher branch.

To further supervise image generation quality, we incorporate a standard denoising loss conditioned on sparse depth for the student network:
\begin{equation}
\mathcal{L}_{p_{\mathcal{S}}} = \mathbb{E}_{{x}_t, t, \tilde{c}, {d}_s, \epsilon \sim \mathcal{N}(0, 1)} \left[\|\epsilon - \tilde{\epsilon}_\theta({x}_t, t, \tilde{c}, {d}_s)\|_2^2\right],
\end{equation}
where ${d}_s$ denotes the diffused latent representation of the sparse depth map $\mathcal{D}_s$. The final objective jointly optimizes distillation and pixel-level supervision:
\begin{equation}
\mathcal{L}_{\mathcal{S}}  = \mathcal{L}_{p_{\mathcal{S}}}  + \alpha \cdot \mathcal{L}_{d},
\end{equation}
where $\alpha$ is a weighting coefficient that controls the relative importance of the distillation loss.
Notably, the proposed distillation is applied only during training and introduces no additional cost at inference.
The fine-tuning process is data-efficient, requiring only few-shot adaptation of ControlNet with less than 10\% of the original training data.

\begin{figure}[t]
  \centering
  \includegraphics[width=1\linewidth]{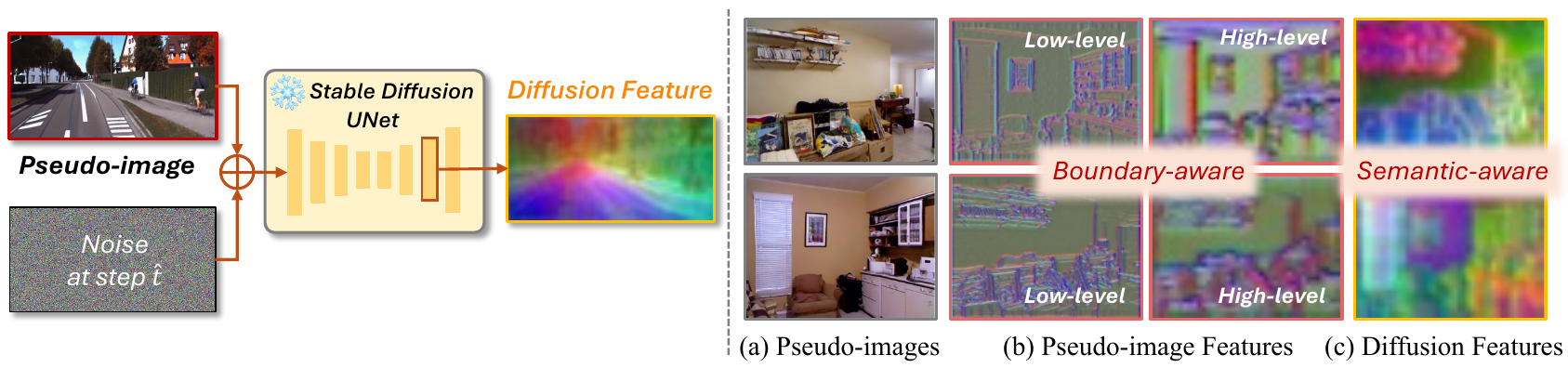}
  % \vspace{-1mm}
  \caption{\textit{Pipeline of our diffusion-based semantic feature extraction and comparative feature analysis}.
The left part illustrates the process of extracting high-level semantic features from the U-Net decoder of Stable Diffusion. The right part compares these diffusion-derived features with pseudo-image features extracted by the image encoder of the guided completion network~\cite{tang2024bilateral}, revealing stronger contextual consistency, richer semantic abstraction, and better structure-awareness in the diffusion-derived representations.
}
\label{stable_diffusion}
\end{figure}

\subsection{Pseudo-image Semantic-guided Completion}
\label{dep_enh}
With the generated pseudo-images, we next focus on effectively integrating their semantic and structural guidance into existing image-guided depth completion methods, forming our generative pseudo-image-guided depth completion framework.
Importantly, this framework is model-agnostic and can be readily incorporated into a broad range of existing guided depth completion architectures.
In the following, we first briefly revisit the conventional image-guided depth completion pipeline.
Building upon it, we introduce our \textit{pseudo-image semantic attention fusion} module, which enables reliable pseudo-image information transfer for enhanced depth estimation via two key components: semantic-guided image feature correction and global image-depth feature interaction.

\begin{wrapfigure}{r}{0.5\textwidth}
\vspace{-3mm}
  \centering
  \includegraphics[width=1\linewidth]{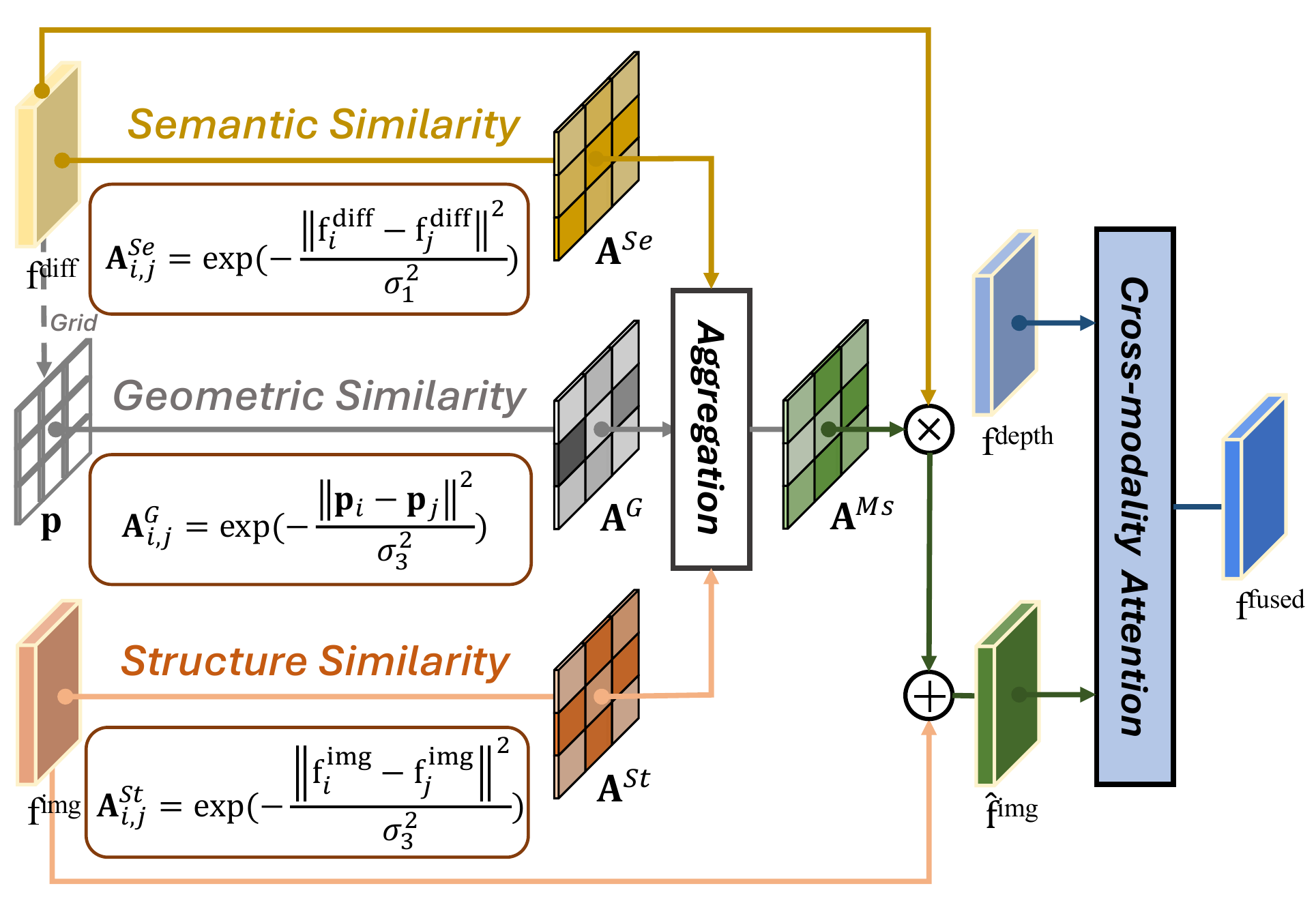}
  % \vspace{1mm}
  \caption{ \textit{Architecture of our Pseudo-image Semantic Attention Fusion module}, which enhances pseudo-image features using semantic cues from diffusion features and adaptively fuses them with depth features for semantics-aware depth completion.
}
\vspace{-3mm}
\label{Attention}
\end{wrapfigure}

\noindent\textbf{Conventional Image-guided Depth Completion.}
Existing guided depth completion networks \cite{shivakumar2019dfusenet, park2020non, li2020multi, tang2024bilateral} typically consist of two independent backbone branches for encoding image features $\mathbf{f}^{\text{img}}$ and depth features $ \mathbf{f}^{\text{depth}}$, respectively.  
Most approaches then directly concatenate them  for semantic-geometric fusion, followed by a decoder to predict the dense depth map.

Unfortunately, in our pseudo-image-guided setting, direct feature concatenation often leads to suboptimal guidance and limited performance gains, mainly due to two factors.
\textbf{i)} \textit{Hallucinations in pseudo-image features:}
Given a sparse depth map as input, the generated pseudo-images may contain hallucinated fine details due to the inherent stochasticity of the generative process, especially in regions with extreme sparsity, thereby degrading the reliability of depth guidance.
\textbf{ii)} \textit{Local feature bias in conventional guided models:}
When pseudo-images are processed by conventional guided completion networks, the extracted features tend to overemphasize local edge details across both low- and high-level layers because of architectural bias toward fine-grained cues over global semantics, as shown in Fig.~\ref{stable_diffusion}$(b)$, which amplifies misleading information and artifacts present in the pseudo-images.

\noindent\textbf{Semantic-guided Image Feature Correction.}
To mitigate feature distortions caused by hallucinated artifacts in pseudo-images, we propose a \textit{semantic-guided image feature correction} module.
Our goal is to mine reliable, high-level semantic information embedded in pseudo-images and use it to correct low-level image feature distortions. Specifically, our correction module complements conventional image encoder features $\mathbf{f}^{\text{img}}$ with powerful semantic representations from pretrained large-scale vision foundation models through a multi-source affinity kernel, thereby producing semantically enhanced and structurally consistent pseudo-image features for depth guidance.

Specifically, we leverage Stable Diffusion to extract semantically rich and contextually aware pseudo-image features, referred to as \textit{diffusion features} $\mathbf{f}^{\text{diff}}$, which provide semantically stable, spatially coherent, and globally informative representations complementary to conventional image encoder features $\mathbf{f}^{\text{img}}$.
Given a generated pseudo-image, we first encode it into the latent space using the pre-trained VAE encoder of Stable Diffusion and then inject Gaussian noise at a predefined timestep $\hat{t}$.
The resulting noisy latent representation is passed through the U-Net backbone (\ie, the denoiser) of Stable Diffusion for progressive feature encoding and decoding.
Among its twelve decoding layers, ranging from high- to low-level semantics, we empirically select the features from the 6\textsuperscript{th} layer as the final pseudo-image diffusion features, as they offer a balanced representation that captures both global contextual semantics and local structural details (see Fig.~\ref{stable_diffusion}$(c)$).

Subsequently, we integrate the diffusion feature with the conventional pseudo-image feature using a multi-source affinity kernel $A^{Ms}$, which jointly models geometric, structural, and semantic consistency between these two complementary feature sources, as demonstrated in Fig.~\ref{Attention}. 
By estimating cross-source feature affinities, this kernel enables selective and reliability-aware aggregation of diffusion features into pseudo-image representations, producing semantically enriched, spatially coherent, and geometry-preserving features for robust depth guidance. 
The enhanced feature $\hat{\mathbf{f}}^{\text{img}}_i$ is defined as:
\begin{equation}
\hat{\mathbf{f}}^{\text{img}}_i = \mathbf{f}^{\text{img}}_i + \sum_{j=1}^{HW} \frac{A^{Ms}_{i,j}}{\sum_{j'=1}^{H \times W} A^{Ms}_{i,j'}}\mathbf{f}^{\text{diff}}_j,
\end{equation}
where $\mathbf{f}^{\text{diff}}_j$ and $\mathbf{f}^{\text{img}}_i$ denote the diffusion feature and pseudo-image feature at positions $j$ and $i$, respectively, with $1\le i,j\le H\times W$. Our multi-source affinity kernel $A^{Ms}$ incorporates three similarity components: structural similarity $A^{St}$, semantic similarity $A^{Se}$, and geometric similarity $A^{G}$, defined by $A^{Ms}_{i,j} = A_{i,j}^{G} \cdot \frac{1}{2} \left( A_{i,j}^{St} + A_{i,j}^{Se} \right)$. These affinity matrices are defined as:
\begin{itemize}
    \item \textbf{Structure similarity.} \( A_{i,j}^{St} \) measures the feature similarity between pseudo-image encoder features $\mathbf{f}^{\text{img}}_i$ and $\mathbf{f}^{\text{img}}_j$, encouraging message passing across regions with similar appearances: 
\begin{equation}\small
    A_{i,j}^{St} = \exp \left( -\frac{\|\mathbf{f}^{\text{img}}_i- \mathbf{f}^{\text{img}}_j\|^2}{\sigma_1^2} \right).
\end{equation}
    \item \textbf{Semantic similarity.} \( A_{i,j}^{Se} \) measures the semantic affinity between diffusion features $\mathbf{f}^{\text{diff}}_i$ and $\mathbf{f}^{\text{diff}}_j$, leveraging their high-level semantic abstraction to capture long-range contextual dependencies and guide robust diffusion feature aggregation:
 \begin{equation}\small
    A_{i,j}^{Se} = \exp \left( -\frac{\|\mathbf{f}^{\text{diff}}_i- \mathbf{f}^{\text{diff}}_j\|^2}{\sigma_2^2} \right).
\end{equation}
    \item \textbf{Geometric similarity.} \( A_{i,j}^{G} \) encodes the spatial proximity between the pixel coordinates \( \mathbf{p}_i, \mathbf{p}_j \in \mathbb{R}^2 \) , enforcing the local spatial coherence:
\begin{equation}\small
    A_{i,j}^{G} = \exp \left( -\frac{\|\mathbf{p}_i - \mathbf{p}_j\|^2}{\sigma_3^2} \right).
\end{equation}
\end{itemize}

\noindent\textbf{Global Image--Depth Feature Interaction.}
Building on the enhanced pseudo-image features $\hat{\mathbf{f}}^{\text{img}}$, instead of directly concatenating them with depth features $\mathbf{f}^{\text{depth}}$ from the depth encoder, we introduce a \textit{global image-depth feature interaction} block to enable adaptive interaction between the visual and geometric modalities. Specifically, we adopt a standard cross-attention mechanism, where the \textit{query} is projected from the depth features $\mathbf{f}^{\text{depth}}$, while the \textit{key} and \textit{value} are projected from the enhanced pseudo-image features $\hat{\mathbf{f}}^{\text{img}}$. This allows each depth token to retrieve relevant semantic and structural cues from the pseudo-image representation. The cross-attention fusion is formulated as:
\begin{equation}
\mathbf{f}^{\text{fused}} =
\mathbf{f}^{\text{depth}}+
\text{Softmax}
\left(
\frac{\mathbf{Q}\mathbf{K}^{\top}}{\sqrt{d}}
\right)
\mathbf{V}.
\end{equation}

The resulting fused feature $\mathbf{f}^{\text{fused}}$ is then fed into the depth completion decoder for final dense depth prediction.

\begin{figure*}[t]
  \centering
  \includegraphics[width=1\linewidth]{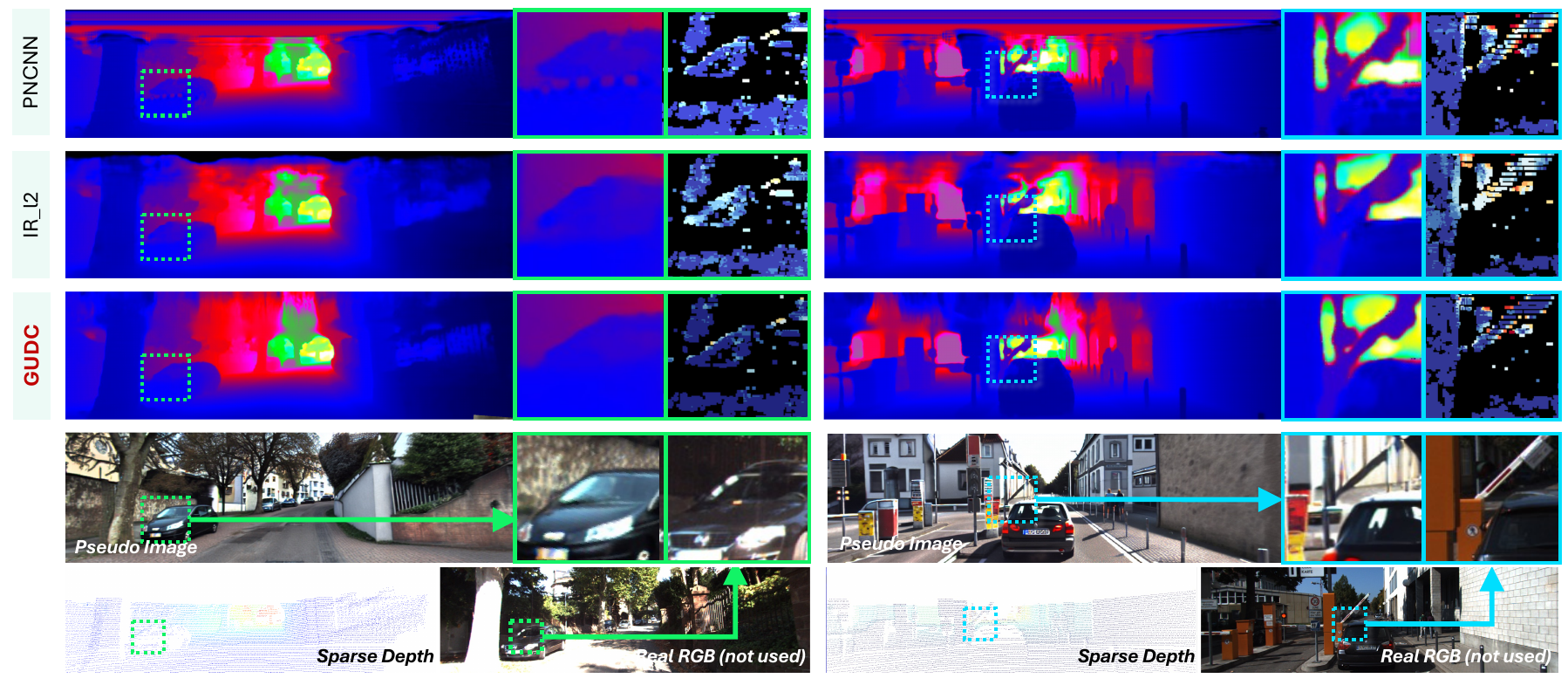}
  \vspace{1mm}
  \caption{Qualitative comparisons with SOTA unguided methods on KITTI~\cite{uhrig2017sparsity} test set, using official visualizations from the benchmark website.
Our generative pseudo-images are shown in the second-to-last row for side-by-side comparison with real RGB images.
}
\label{KITTI_results}
\end{figure*}

\section{Experiment}
We evaluate our Generative Unguided Depth Completion (GUDC) framework from two key aspects: \textbf{i)} completion accuracy in unguided settings, and \textbf{ii)} robustness under adverse visual conditions where guided methods typically fail.

\subsection{Experimental Setup}
\noindent\textbf{Datasets.} 
Experiments are conducted on two standard benchmarks: KITTI~\cite{uhrig2017sparsity} (outdoor LiDAR-based driving scenarios) and NYUv2~\cite{silberman2012indoor} (indoor RGB-D scenes captured by Kinect).
For KITTI, we follow the official split with the default sparse depth input containing approximately 5\% valid pixels.
For NYUv2, sparsity is simulated by randomly sampling 200 and 500 depth points per frame, consistent with standard practice.

\begin{wraptable}{r}{0.5\textwidth}
  \vspace{-4mm}
\small
\centering
\resizebox{1\linewidth}{!}{
\begin{tabular}{lcccc}
\toprule
\textbf{Methods} & RMSE $\downarrow$ & MAE $\downarrow$ & iRMSE $\downarrow$ & iMAE $\downarrow$ \\
\midrule
SI-CNN~\cite{uhrig2017sparsity}        & 1601.33 & 481.27 & 4.94  & 1.78 \\
DCCS~\cite{chodosh2019deep}      & 1325.37 & 439.48 & 59.39 & 3.19 \\
NConv-CNN~\cite{eldesokey2018propagating}  & 1268.22 & 360.28 & 4.67  & 1.52 \\
Spade-sD~\cite{jaritz2018sparse}    & 1035.29 & 248.32 &2.60 & 0.98 \\
PNCNN~\cite{eldesokey2020uncertainty}       & 960.05  & 251.77 & 3.37  & 1.05 \\
S$^3$2D$_d$~\cite{ma2019self}       & 954.36  & 288.64 & 3.21  & 1.35 \\
HMS-Net~\cite{huang2019hms}       & 937.48  & 258.48 & 2.93  & 1.14 \\
Glob\_guide~\cite{van2019sparse}   & 922.93  & 249.11 & 2.80  & 1.07 \\
IR\_l1~\cite{lu2020depth} & 915.86  & 231.37 & 3.19  & 1.23 \\
IR\_l2~\cite{lu2020depth} & 901.43 & 292.36 & 4.92  & 1.35 \\
DCAE~\cite{lu2022depth}    & 1451.67 & 429.74 & 4.89  & 1.78 \\
\midrule
\textbf{GUDC (Ours)}& \textbf{882.31} & \textbf{228.22} & \textbf{2.44} & \textbf{0.94} \\ 
\bottomrule
\end{tabular}
}
\caption{Quantitative comparisons with SOTA unguided methods on KITTI~\cite{uhrig2017sparsity}. 
}
\label{tab1_kitti}
 \vspace{2mm}
\centering
\resizebox{1\linewidth}{!}{ 
  \begin{tabular}{l cc cc}
    \toprule
    \textbf{Methods} & 
    \multicolumn{2}{c}{\textbf{200 samples}} & 
    \multicolumn{2}{c}{\textbf{500 samples}} \\
    & RMSE $\downarrow$ & REL $\downarrow$ 
    & RMSE $\downarrow$ & REL $\downarrow$ 
    \\
    \midrule
    SparseConv~\cite{uhrig2017sparsity} & 1.065 & 0.257 & 0.801 & 0.159 \\
    S2D~\cite{ma2018sparse} & 0.259 & 0.054 & 0.230 & 0.044 \\
    MS-IDWNet~\cite{mengistu2023embedding} &  0.255 & 0.055 & 0.190 & 0.038 \\
     EIR-Net~\cite{wei2022efficient} &  - & - & 0.142 & 0.020 \\
SIUNet~\cite{ramesh2023siunet}  & 0.204 &  0.033 & 0.138 & 
     0.015 \\
BUNet~\cite{liu2024towards} &  0.195 & 0.030 & - & - \\
    \midrule
    \textbf{GUDC (Ours)}
                & \textbf{0.189} & \textbf{0.028} & \textbf{0.133} & \textbf{0.014}  \\
    \bottomrule
  \end{tabular}
}
\caption{Quantitative comparisons with SOTA unguided methods on NYUv2~\cite{silberman2012indoor}.}
\label{tab_NYU_results}
 \vspace{2mm}
\centering
\resizebox{1\linewidth}{!}{ 
  \begin{tabular}{l cc cc}
    \toprule
    \textbf{Methods} & 
    \multicolumn{2}{c}{\textbf{Foggy-KITTI}} & 
    \multicolumn{2}{c}{\textbf{Nighttime-KITTI}} \\
    & RMSE $\downarrow$ & MAE $\downarrow$ 
    & RMSE $\downarrow$ & MAE $\downarrow$ 
    \\
    \midrule
    S$^3$2D$_{rgbd}$~\cite{ma2019self}  & 1115.34 & 309.65 & 1244.83 & 355.11 \\
    GuideNet~\cite{tang2020learning} & 960.23 & 251.24 & 971.81 & 293.29 \\
    BP-Net~\cite{tang2024bilateral} & 952.34 & 243.35 & 972.86 & 276.29 \\
    DFU-Net~\cite{wang2024improving} & 967.84 & 289.29 & 984.53 & 294.34 \\
    \midrule
    \textbf{GUDC (Ours)}
                & \textbf{932.24} & \textbf{235.32} & \textbf{932.24} & \textbf{235.32}  \\
    \bottomrule
  \end{tabular}
}
 \caption{Quantitative comparisons with SOTA image-guided methods on Foggy-KITTI and Nighttime-KITTI.}
\vspace{2mm}
\label{Foggy_KITTI_tab}
\end{wraptable}

To assess robustness under degraded visual conditions, we construct two adverse-condition variants: Foggy-KITTI, synthesized via an atmospheric scattering model~\cite{narasimhan2002vision}, and Nighttime-KITTI, generated using the image-to-image translation model img2img-Turbo~\cite{parmar2024one}.

\noindent\textbf{Implementation Details.}
During pseudo-image generation, we randomly sample 6K and 5K RGB-depth pairs from KITTI and NYUv2 (less than 10\%) for few-shot ControlNet fine-tuning. 
Following the default settings in ControlNet~\cite{zhang2023adding}, we use AdamW with a learning rate of 1e-5, training for 11 epochs with a distillation weight of $\alpha{=}1$.
For depth completion, we adopt BP-Net~\cite{tang2024bilateral} as the backbone and replace its image-depth fusion module with our semantic attention fusion module, setting $\sigma_1{=}\sigma_2{=}0.5$ and $\sigma_3{=}0.1$. The network is fine-tuned on less than 30\% of the available training data (10K samples for NYUv2 and 20K for KITTI), with the diffusion timestep set to $\hat{t}{=}20$.
All experiments are conducted on NVIDIA RTX 4090 and TITAN RTX.

We compare GUDC against four state-of-the-art (SOTA) guided methods (BP-Net~\cite{tang2024bilateral}, S$^3$2D$_{rgbd}$~\cite{ma2019self}, GuidedNet~\cite{tang2020learning},  DFU-Net~\cite{wang2024improving}) and seventeen SOTA unguided methods, including sparsity-aware CNNs (SI-CNN~\cite{uhrig2017sparsity}, SparseConv~\cite{uhrig2017sparsity}, Spade-sD~\cite{jaritz2018sparse}, DCCS~\cite{chodosh2019deep}, HMS-Net~\cite{huang2019hms}, EIR-Net~\cite{wei2022efficient}, MS-IDWNet~\cite{mengistu2023embedding}), normalized convolution models (NConv-CNN~\cite{eldesokey2018propagating}, PNCNN~\cite{eldesokey2020uncertainty}), standard CNN baselines (S2D~\cite{ma2018sparse}, S$^3$2D$_d$~\cite{ma2019self}, Glob\_guide~\cite{van2019sparse}), and implicit visual-based networks (DCAE~\cite{lu2022depth}, IR\_l1/ IR\_l2~\cite{lu2020depth}, SIUNet~\cite{ramesh2023siunet}, BUNet~\cite{liu2024towards}).

\subsection{Comparison with SOTAs}
\noindent\textbf{Evaluation on KITTI.} We first perform model evaluation on a widely-used outdoor benchmark dataset, KITTI~\cite{uhrig2017sparsity}. Table~\ref{tab1_kitti} presents the quantitative comparison on KITTI test set under standard sparse depth settings. 
Our method achieves the best performance across all evaluation metrics, consistently outperforming both purely depth-based approaches and implicit visual-supervised methods by a significant margin. 
These results demonstrate the effectiveness of our generative guidance paradigm, which introduces explicit visual priors during inference to enhance structural recovery and fine-grained depth completion.

\begin{wrapfigure}{r}{0.5\textwidth}
 \centering
  \includegraphics[width=1\linewidth]
  {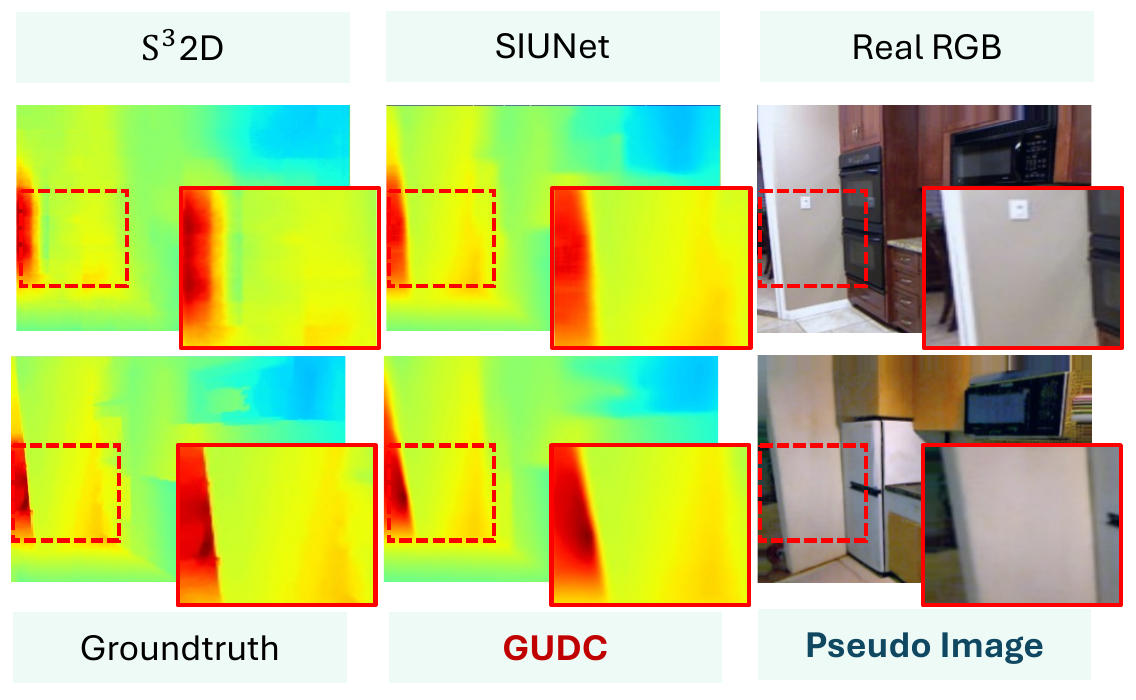}
  \caption{Qualitative comparisons with SOTA methods on NYUv2~\cite{silberman2012indoor} (200 samples).
}
\label{NYU_visual_results}
  \centering
  \includegraphics[width=1\linewidth]{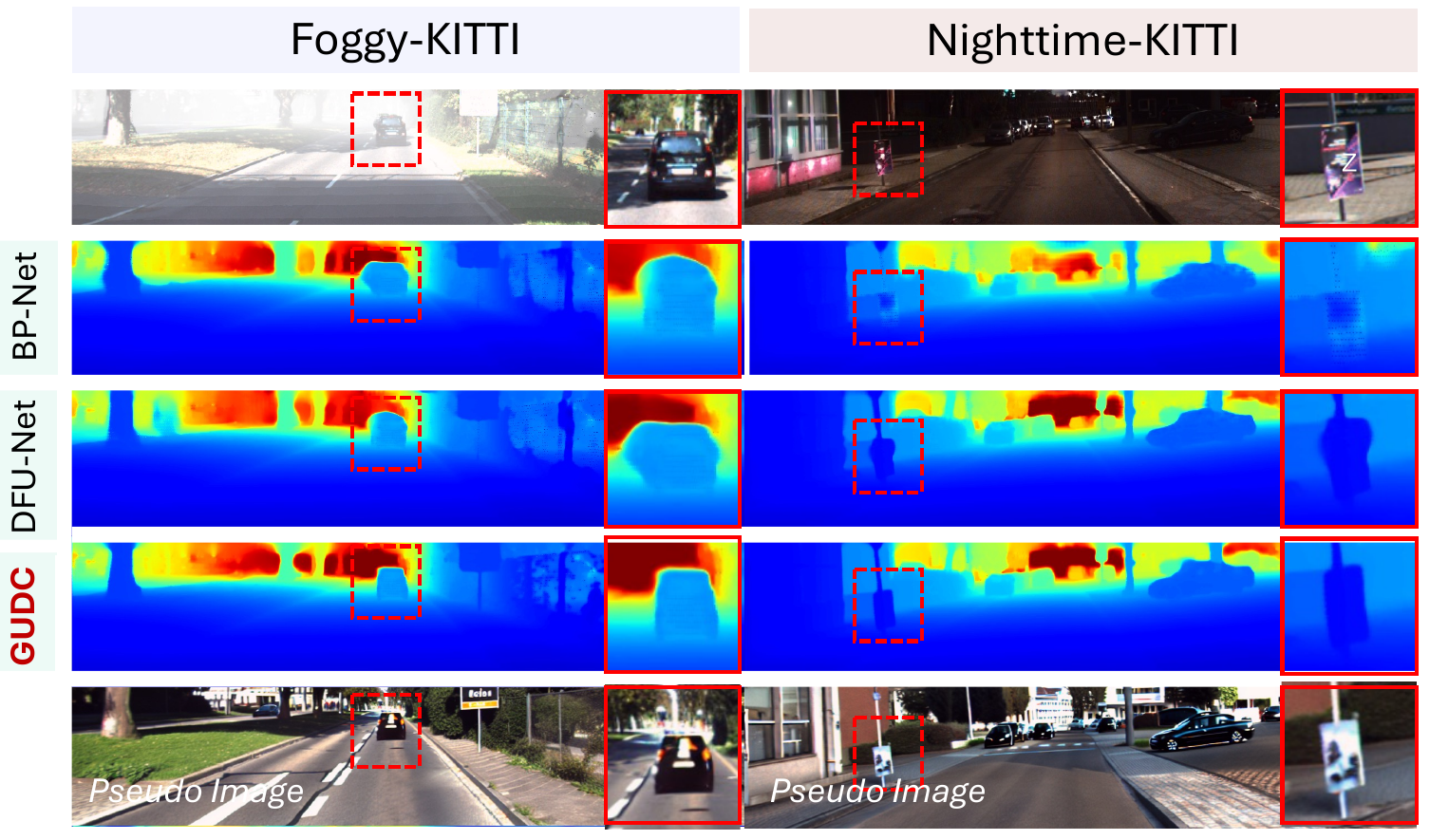}
  \caption{Qualitative comparisons with image-guided depth completion methods on Foggy-KITTI and Nighttime-KITTI datasets.
}
\label{Foggy_KITTI}
\end{wrapfigure}

Fig.~\ref{KITTI_results} provides qualitative comparisons with state-of-the-art (SOTA) methods on KITTI test set. 
For fair comparison, we utilize the official visual results released on the KITTI benchmark website for all competing methods.  
Our predictions exhibit noticeably sharper object boundaries and fewer artifacts in severely sparse regions, where other methods tend to produce blurry or incomplete estimations. 
These improvements stem from our high-quality and structurally consistent pseudo-RGB images, which provide explicit semantic and contextual cues during inference. 
Such visual priors effectively guide the network toward more accurate depth recovery, particularly in regions with missing or ambiguous depth input.

\noindent\textbf{Evaluation on NYUv2.}
We further assess GUDC against state-of-the-art unguided methods on the NYUv2 indoor benchmark~\cite{silberman2012indoor}.
Table~\ref{tab_NYU_results} summarizes the quantitative results under varying levels of input sparsity (200 and 500 points per frame).  
Our method consistently outperforms all prior unguided approaches across all sparsity levels, achieving the lowest RMSE and REL.  
Qualitative comparisons on NYUv2 are shown in Fig.~\ref{NYU_visual_results}. Our method produces sharper depth boundaries and more accurate reconstructions of fine-scale indoor structures (\eg, furniture edges and room layouts), while competing methods often yield over-smoothed or structurally distorted outputs. These improvements validate the effectiveness of introducing strong visual semantics via our Pseudo-image Semantic-Attention Fusion module.

\noindent\textbf{Evaluation on Foggy-KITTI and Nighttime-KITTI.} 
To assess the robustness of GUDC under degraded visual conditions, we further compare it with representative image-guided depth completion baselines on two challenging KITTI~\cite{uhrig2017sparsity} variants: Foggy-KITTI and Nighttime-KITTI. These two settings simulate two common real-world failure cases for RGB-guided completion, where visual observations are severely corrupted by atmospheric scattering or low illumination. As shown in Table~\ref{Foggy_KITTI_tab}, conventional image-guided methods suffer noticeable performance degradation under heavy fog or nighttime conditions, since their RGB branches tend to extract unreliable edges, weakened textures, and corrupted semantic cues. Such degraded image features are then propagated into the depth branch, leading to inaccurate depth recovery and unstable predictions.

\begin{wraptable}{r}{0.5\textwidth}
\vspace{-2mm}
\resizebox{1\linewidth}{!}{
\small
\begin{tabular}{l c}
\toprule
\textbf{Method} & \textbf{RMSE} $\downarrow$ \\
\midrule
\multicolumn{2}{l}{\textit{Effect of Multi-level Dense-to-sparse Distillation}} \\
\cmidrule(lr){1-2}
Finetune\_ControlNet w/o Distillation Loss & 941.13 \\
Finetune\_ControlNet (\#L=1) & 934.11 \\
Finetune\_ControlNet (\#L=2) $^{\ast}$ & \textbf{932.24} \\
Finetune\_ControlNet (\#L=3) & 932.87 \\
\midrule
\multicolumn{2}{l}{\textit{Effect of ControlNet Fine-Tuning Sample Size}}\\
\cmidrule(lr){1-2}
Finetune\_ControlNet (\#samples=1K) & 941.32 \\
Finetune\_ControlNet (\#samples=4K) & 936.75 \\
Finetune\_ControlNet (\#samples=6K) $^{\ast}$ & 932.24 \\
Finetune\_ControlNet (\#samples=8K) & \textbf{931.65}\\
\midrule
\multicolumn{2}{l}{\textit{Effect of PSAF and Its Components}} \\
\cmidrule(lr){1-2}
Full GUDC $^{\ast}$ & \textbf{932.24}\\
w/o PSAF & 943.53 \\
w/o Image Feature Correction & 937.45 \\
w/o Global Feature Interaction &  940.75\\
\midrule
\multicolumn{2}{l}{\textit{Compatibility with Various Depth Completion Backbones}} \\
\cmidrule(lr){1-2}
BP-Net (depth-only) & 953.34 \\
BP-Net + \textbf{GUDC} (Ours) $^{\ast}$ & \textbf{932.24}\\
\hline
MSG-CHN (depth-only) & 1004.05  \\
MSG-CHN + \textbf{GUDC} (Ours)& \textbf{974.67}\\
\hline
DFU-Net (depth-only) & 957.54 \\
DFU-Net + \textbf{GUDC} (Ours)& \textbf{939.23} \\
\bottomrule
\end{tabular}
}
\caption{Ablation studies on KITTI validation dataset. ($\ast$) denotes the default configuration.}
\label{tab_ab}
\end{wraptable} 

In contrast, our GUDC consistently maintains high accuracy across both adverse settings and outperforms all guided baselines by a clear margin. This robustness mainly stems from our unguided generative formulation: instead of directly relying on the corrupted real RGB images, GUDC synthesizes geometry-aligned pseudo-images from the sparse depth input and uses them as semantic priors for depth completion. Since the pseudo-images are conditioned on depth rather than degraded visual observations, they provide cleaner structural guidance and more stable contextual cues under challenging illumination and weather conditions. Qualitative comparisons in Fig.~\ref{Foggy_KITTI} further verify this advantage. While guided methods produce blurred boundaries, distorted object structures, or inconsistent depth in visually corrupted regions, GUDC recovers sharper structures and more coherent depth maps. These results demonstrate that removing the dependence on real RGB inputs enables GUDC to achieve superior robustness under extreme visual degradation.

\subsection{Ablation Study}

\begin{figure}[t]
  \centering
  \includegraphics[width=0.98\linewidth]{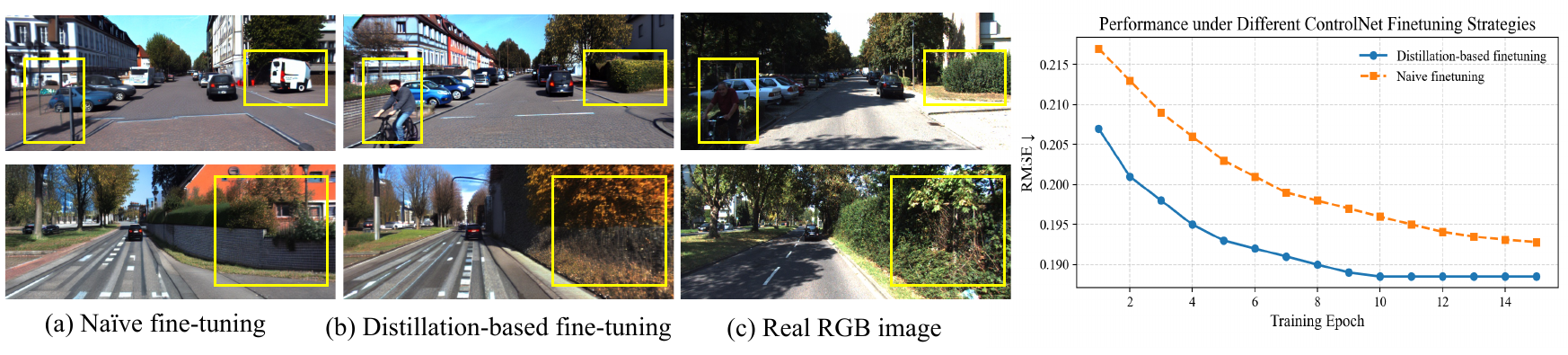}
  \caption{\textit{Left}: Qualitative comparisons of pseudo-images generated by naïve fine-tuning and our distillation-based fine-tuning on KITTI. \textit{Right}: Epoch-wise performance on NYUv2 test set for depth completion guided by both types of pseudo-images.
}
\label{ab_results}
\end{figure}

\noindent\textbf{Effect of Distillation-based ControlNet Fine-Tuning.}
We first evaluate dense-to-sparse distillation for fine-tuning ControlNet. As shown in Fig.~\ref{ab_results}, naïve fine-tuning suffers from hallucinations and geometric misalignment in sparsely observed regions, whereas our distillation-based strategy produces pseudo-images with clearer geometry and better depth-layout consistency. The epoch-wise RMSE curves further show faster convergence and improved performance, confirming that our distillation improves pseudo-image guidance.
We further conduct quantitative ablations in Table~\ref{tab_ab} (\textit{first block}). Removing $\mathcal{L}_d$ causes clear performance degradation, confirming that teacher-guided representation alignment is essential for reliable generation. Varying the number of aligned layers $L$ shows that multi-level distillation improves over single-layer alignment, while increasing $L$ from 2 to 3 yields only marginal gains. Thus, we adopt $L=2$ as the default setting for a favorable accuracy-efficiency trade-off.

Finally, we study the sample efficiency of ControlNet fine-tuning. As shown in Table~\ref{tab_ab} (\textit{second block}), performance improves as the amount of training data increases, but gradually saturates around 6K samples, with only a negligible gain when increasing the training set to 8K samples. We therefore use 6K samples as the default configuration. Notably, even with only 1K samples, the fine-tuned ControlNet achieves satisfactory performance, demonstrating that our adaptation is data-efficient and can effectively exploit limited supervision.

\noindent\textbf{Ablation of Our PSAF module.} 
We evaluate the effectiveness of our Pseudo-image Semantic Attention Fusion (PSAF) module and its components via an ablation study involving four variants:  
(1) full PSAF,  
(2) direct concatenation of pseudo-image and depth features without PSAF,  
(3) PSAF without the Semantic-Guided Image Feature Correction Unit, and  
(4) PSAF without the Global Image-Depth Feature Interaction Block.
As shown in Table~\ref{tab_ab} (\textit{third block}), removing either component leads to noticeable performance drops. The full PSAF achieves the best overall results, confirming the benefits of image feature refinement and adaptive cross-modal interaction for accurate and robust depth completion.

\noindent\textbf{Compatibility with Various Depth Completion Backbones.} To verify the generality and flexibility of our proposed framework, we evaluate its integration with different depth completion backbones. 
Specifically, we compare the performance of our generative completion framework when applied to three representative models:  
(1) BP-Net~\cite{tang2024bilateral}, a standard encoder-decoder architecture with image-depth fusion;  
(2) MSG-CHN~\cite{li2020multi}, a cascaded hourglass network with one image branch and three scale-specific depth branches; and    
(3) DFUNet~\cite{wang2024improving}, a dual-branch guided completion network with hierarchical fusion.
As shown in Table~\ref{tab_ab} (\textit{fourth block}), our generative paradigm can serve as a plug-and-play enhancement across a variety of depth completion frameworks.

\section{Conclusion}
In this work, we proposed a novel paradigm termed \textit{Generative Unguided Depth Completion} for unguided depth completion, which bridges generative models with unguided depth completion to enhance depth precision without requiring real RGB guidance. By leveraging depth-conditioned ControlNet, we generated geometrically aligned pseudo-images that served as structural and semantic guidance. To address the misalignment caused by depth sparsity, we proposed a multi-level dense-to-sparse representation distillation strategy to boost image quality. Furthermore, we introduced the Pseudo-image Semantic-Attention Fusion module to extract reliable semantic cues while suppressing hallucinated artifacts. Extensive experiments on the KITTI and NYUv2 datasets demonstrated the effectiveness and generalization capability of our method.

\bibliography{egbib}
\end{document}